\documentclass{article}

\usepackage[final]{neurips_ICBINBW}

\usepackage[utf8]{inputenc} 
\usepackage[T1]{fontenc}    
\usepackage{hyperref}       
\usepackage{url}            
\usepackage{booktabs}       
\usepackage{amsfonts}       
\usepackage{nicefrac}       
\usepackage{microtype}      
\usepackage{xcolor}         

\usepackage{subfigure}
\usepackage{caption}
\usepackage{amsmath}
\usepackage{graphicx}

\title{Exploring the Long-Term Generalization \\ of Counting Behavior in RNNs}

\author{
   Nadine El-Naggar  \hspace{2em}
   Pranava Madhyastha \hspace{2em}  Tillman Weyde \\
    City, University of London \\ United Kingdom \\
   \texttt{\{nadine.el-naggar,pranava.madhyastha,t.e.weyde\}@city.ac.uk} 
 }

\begin{document}

\maketitle

\begin{abstract}
In this study, we investigate the  generalization of LSTM, ReLU and GRU models on counting tasks over long sequences. 
Previous theoretical work has established that RNNs with ReLU activation and LSTMs have the capacity for counting with suitable configuration, while GRUs have limitations that prevent correct counting over longer sequences.  
Despite this and some positive empirical results for LSTMs on Dyck-1 languages, our experimental results show that LSTMs fail to learn correct counting behavior for sequences that are significantly longer than in the training data. 
ReLUs show much larger variance in behavior and in most cases worse generalization.
The long sequence generalization is empirically related to validation loss, but reliable long sequence generalization seems not practically achievable through backpropagation  with current techniques. 
We demonstrate different failure modes for LSTMs, GRUs and ReLUs. 
In particular, we observe that the saturation of activation functions in LSTMs and the correct weight setting for ReLUs to generalize counting behavior are not achieved in standard training regimens.  
In summary, learning generalizable counting behavior is still an open problem and we discuss potential approaches for further research. 
\end{abstract}

\section{Introduction}

Recurrent Neural Networks (RNNs) have been the go-to neural model for sequential tasks since the 1980s and have been popular in recent years for a variety of applications, \citep[see e.g., ][]{karpathy2015unreasonable}, including Natural Language Processing (NLP). 
RNNs are highly expressive, as they are theoretically Turing complete \citep{DBLP:journals/jcss/SiegelmannS95} and inherit universal approximation capacity \citep{funahashi1992neural,leshno1993multilayer} from multi-layer perceptrons.
Although RNNs have largely been superseded by Transformers \citep{vaswani2017attention} in NLP, there is recently a renewed interest in their theoretical properties and specifically the counting behavior 
\citep{DBLP:conf/acl/MerrillWGSSY20,srivastava2022beyond}.

Recent work by 
\citet{suzgun2019lstm} and 
\citet{Weiss-et-al-18} 
follows in the footsteps of \citet{DBLP:journals/tnn/GersS01} to evaluate the ability of RNN models to learn formal languages in simplified scenarios. 
\citet{Weiss-et-al-18}
show that 
single-cell Long Short-Term Memory (LSTM) and Rectified Linear Unit (ReLU) networks
have the capacity for unbounded counting and modeling a set of formal languages (using saturation of some activation functions in LSTMs), while 
Gated Recurrent Units (GRUs)
and Elman RNNs do not. 
\citet{Weiss-et-al-18} also
report that ReLUs are difficult to train,
which is consistent with the findings by \cite{hochreiter1991untersuchungen} and \cite{bengio1994learning} that RNNs suffer from vanishing and exploding gradients when trained on tasks with long term dependencies.
LSTM, introduced by \citet{DBLP:journals/neco/HochreiterS97}, remedies some of these problems.

\citet{suzgun2019lstm} show empirically that single-cell LSTMs can learn counting behavior to recognize Dyck-1 languages with 100\% accuracy in their experiments. 
However, several open questions remain, including whether the counting exhibited by LSTMs is sufficiently precise for long sequences and to what extent ReLUs and GRUs can learn counting on long sequences. 
In this work, we investigate the limits of trained ReLU, LSTM and GRU models in generalizing Dyck-1 recognition to sequences that are substantially longer than those in the training set and analyze when, how and why they fail. 
%
The specific contributions of this study are:
\begin{enumerate}
    \item We observe that single-cell LSTM, ReLU and GRU models generally fail to accept Dyck-1 sequences of length 1000 when trained on lengths up to 50.
    \item We further observe that validation loss is a good  predictor for long sequence generalization, but reducing loss through longer training is not a practical method to achieve better generalization.
    \item We illustrate and analyze the different failure modes of ReLU, LSTM, and GRU cells.  
\end{enumerate}
Our code is publicly available at \url{https://github.com/nadineelnaggar/Long\_Term\_Counting\_RNNs}.

\section{Modeling Dyck-1 Sequences}

We evaluate the limit of three different single-cell RNN models trained on Dyck-1 sequences in generalizing to longer sequences. 
Dyck-1 sequences have a) one type of brackets, b) an equal number of opening and closing brackets, and c) no closing brackets before their corresponding opening brackets.
We use the same task setup as \citet{suzgun2019lstm} to evaluate our models, which is to classify which tokens are valid to follow in a Dyck-1 sequence after each token.
This is equivalent to classifying complete vs incomplete Dyck-1 sequences.  
This is limited as a Dyck-1 language acceptance task. 
While the sequence is classified at every time step, sequences that violate condition c)  do not occur. 
Therefore, this setup does not allow for testing full language acceptance, as defined, e.g., in \citet{DBLP:journals/corr/abs-2004-06866}.

For training and initial testing, we use the same types of datasets as reported in \citet{suzgun2019lstm}. 
Our \emph{Training Set} consists of 10000 sequences with lengths of 2 to 50 tokens. 
We use a \emph{Validation Set} with 5000 sequences of length 2 to 50 tokens.
We use a \emph{Long Test Set} with 5000 sequences of lengths 52 to 100 tokens. 
The disjoint Training, Validation and Long Test Sets are generated in the same way as in \citet{suzgun2019lstm}. 

We use single-layer single-cell LSTMs, ReLUs and GRUs in our experiments.
We include GRUs in these experiments, although they cannot perform unbounded counting, to compare how this limitation manifests itself empirically. 
We train online with the Adam optimizer \citep{kingma2014adam}, and learning rates of 0.01 for the ReLU and LSTM models, and 0.001 for the GRU models. 
The training was run using PyTorch in a Linux environment. 

We train 10 models of each type for 30 epochs and select the best model (out of the epochs) per run by validation loss. 
We observe convergence typically within 20 epochs of training, often less.
We experimented with training for 100 epochs on a subset without observing any additional benefit.
For ReLUs we train 30 models and select the 10 best runs, as many ReLUs do not learn to recognize Dyck-1 sequences at all.

\section{Very Long Sequence Generalization}


We present the prediction perfomance results of our experiments in Table \ref{tab:results_accuracy}. We observe that all three models perform well on both Training and Validation Sets, but there is more variation on the Long Test Set.
We also observe in Table~\ref{tab:results_accuracy} that ReLUs tend to show very variable performance. 
LSTMs perform best on average, and GRUs are on average between LSTMs and ReLUs.

We evaluate the generalization limits of our models that train well on short sequences and generalize well to some longer sequences up to 100 tokens. 
We use a \emph{Very Long Test Set} with 100 sequences of length 1000 tokens,  which are generated using same methods as presented in \citet{suzgun2019lstm}.

%

The results on the Very Long Test Set are presented in the last column of Table~\ref{tab:results_accuracy} and are shown as First Point of Failure because all models but one fail on these sequences. 
Only one ReLU model has an 
accuracy of 4.0\% while all the other ReLUs, the LSTMs and the GRUs do not recognize a single Very Long sequence correctly.
Previous work from \citet{Weiss-et-al-18} shows that it is not possible for GRUs to perform unbounded counting, so  it is not surprising that their performance deteriorates sharply for the longer sequences. 
ReLUs and LSTMs do have the capacity for unbounded counting, but we observe that training dynamics does not lead to models that count precisely enough in the long run. 
Interestingly, the only model that recognizes at least some of the Very Long sequences correctly is a ReLU model, despite their lower average performance. 
We will discuss the reasons for this behavior below. 

\begin{table}
\centering
\caption{
For Training, Validation, and Long Test Sets we report the  accuracy in percentage of sequences, where a sequence counts as correctly recognized when all tokens are correctly classified.
The numbers in the table are the averages and (minimum / maximum) over 10 models.  
For the Very Long sequences, we report the First Point of Failure (FPF). 
A maximum FPF value \emph{none} means that the best model(s) did not fail. 
The best results per column are highlighted in bold font.
}
\label{tab:results_accuracy}
\vspace{3mm}
\begin{tabular}{llllll}
\hline
 \textbf{Model} & \textbf{Training}  & \textbf{Validation} & \textbf{Long}   &  \textbf{Very Long (FPF)}\\
\hline
LSTM & \textbf{100} (100 / 100) & \textbf{100} (100 / 100)  & 
\textbf{100} (99.8 / 100)
& \textbf{916.8} (802.4 / 946.4)  \\
ReLU & 97.9 (90.9 / 100) & 97.5 (89.3 / 100) & 74.2 (33.5 / 100) & 871.7 (543.1 / \emph{\textbf{none}} )\\
GRU & 100 (100 / 100) & 100 (100 / 100) & 92.5 (90.72 / 95.6)   & 472.5 (434.7 / 545.2) \\
\hline
\end{tabular}
\end{table}

This result does qualify the statement by \citet{suzgun2019lstm} that LSTMs can perform dynamic counting, insofar as the counting does not generalize indefinitely. 
This is actually not surprising given the results by \citet{Weiss-et-al-18}, where LSTMs fail to generalize to very long sequences on a related task, which can be explained by counting not being precise enough, so that the errors accumulate.


\section{Loss and Generalization}

\begin{table}[b]
\centering
\caption{
Linear regression between the loss values on the different datasets (Training, Validation and Long Test Sets) and the FPF on the Very Long Test Set with $R^{2}$ and p values. }
\label{tab:results_linear_regression}

\vspace{3mm}
\begin{tabular}{llll}
\hline
 &  \textbf{Training} & \textbf{Validation} & \textbf{Long}\\
 \textbf{Model} & $R^{2}$ / p & $R^{2}$ / p & $R^{2}$ / p \\
\hline
LSTM & 0.434 / $1.06\times 10^{-8}$ &0.511 / $1.412\times 10^{-10}$ & 0.593 / $6.36\times 10^{-13}$\\
ReLU & 0.002 / 0.75 & 0.094 / 0.017 & 0.201 / 0.0003 \\
GRU & 0.363 / $3.55\times 10^{-7}$ & 0.218 / 0.00017 & 0.075 / 0.035\\
\hline
\end{tabular}

\end{table}

\begin{figure}[h]
    \centering
    \subfigure[LSTM]{\includegraphics[width=0.32\linewidth]{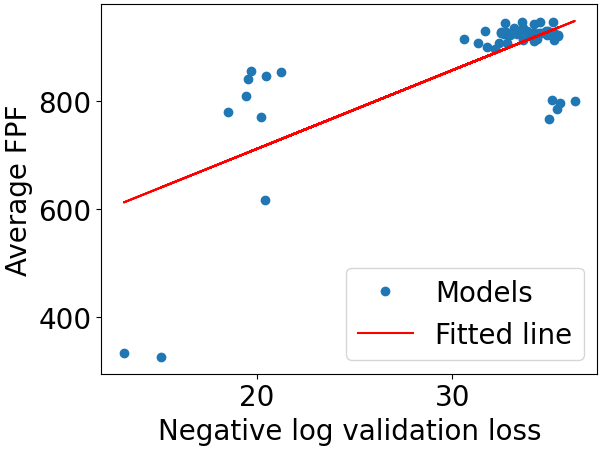}}
    \subfigure[ReLU]{\includegraphics[width=0.32\linewidth]{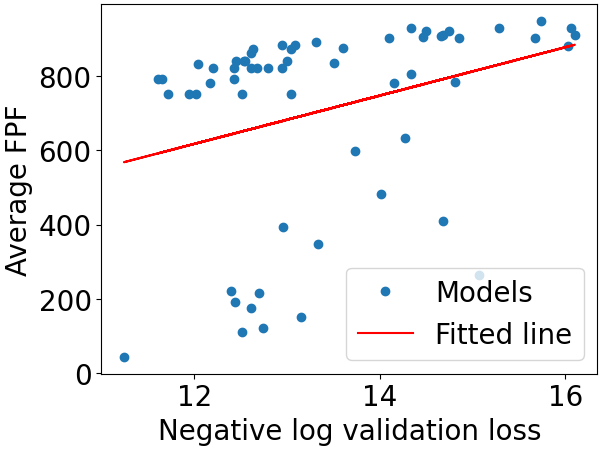}}
    \subfigure[GRU]{\includegraphics[width=0.32\linewidth]{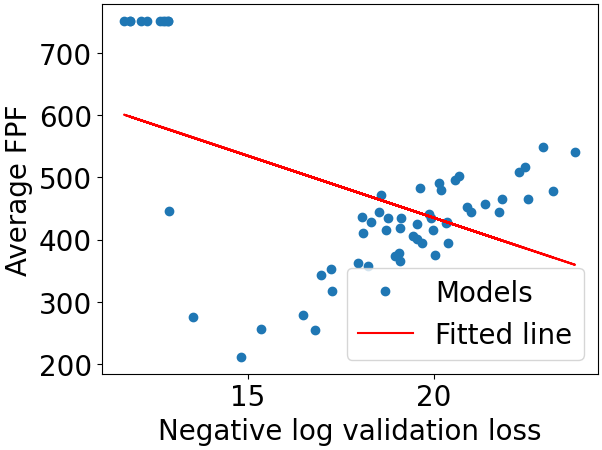}}
    \caption{Scatter plot with regression lines of negative log of the average validation loss versus the average FPF on the Very Long Test Set. 
    We use a set of 60 models from different stages of the training process of for each of ReLU, LSTM and GRU (see text for details). 
    }
    \label{fig:plots_Correlation_zigzag}
\end{figure}

In practice, finite counting abilities may be sufficient, hence it would be useful if it is possible during training (through either the training or validation dynamics) to gather sufficient signals in order to understand how well a model will generalize. 
This is because repeatedly testing on very long sequences can be expensive. In this section, we study the utility of training dynamics towards answering the above question. 
We use the observation that lower loss values achieved by the models seems to be related to higher FPF values. 
We quantify this in a linear regression between the negative log loss on Training, Validation and Long Test Sets and the FPF on the on the Very Long Test Set.
We use here a different set of models, to include different stages of training with different levels of validation loss.  
We take from each of the 10 runs the models after epoch 1, 5, 10,  15, 20, and 25, so that we have 60 models for every type. 
This is done in order to have a range of models including some where the loss values are relatively high because the training was still at the beginning. 
The $R^{2}$ and p values are reported in Table~\ref{tab:results_linear_regression}.

We observe that the losses and FPF are well correlated for the LSTMs. 
For the ReLUs, we see increased correlation for the losses on datasets with long sequences, but overall the correlation is low. 
The GRUs show good correlation on the Training and Validation Sets, but lower correlation values for the Long Test Set, which is plausible given the lower performance on the Long Test Set.  

Figure~\ref{fig:plots_Correlation_zigzag} shows scatter plots with the regression lines for the Validation Set loss. 
For the LSTMs we see that the lower loss (higher negative log loss) does lead to higher FPF, but there are still many values that are 20-25\% lower than the length of the sequences. The ReLUs are much more variable, so that a low validation loss does not guarantee a high FPF. 
The GRUs, are interesting, where we observe that there are a few models with good FPF but relatively high loss. 
These models lead to a negative correlation overall and even for very low losses the FPF values are relatively low.

\subsection{Effect of Training Duration on Loss}
\begin{figure}[h!]
    \centering
    \subfigure[LSTM]{\includegraphics[width=0.491\textwidth]{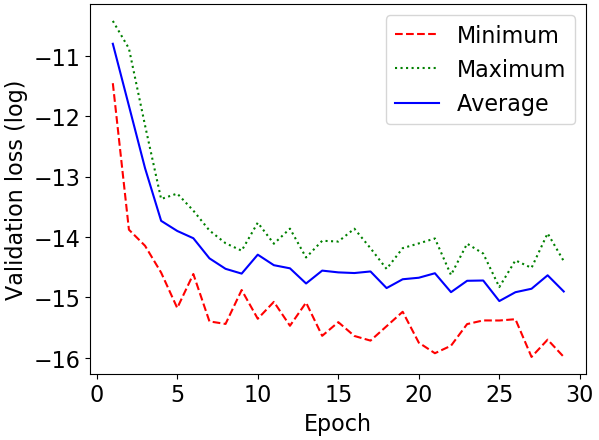}}
    \hspace{3mm}
    \subfigure[ReLU]{\includegraphics[width=0.473 \textwidth]{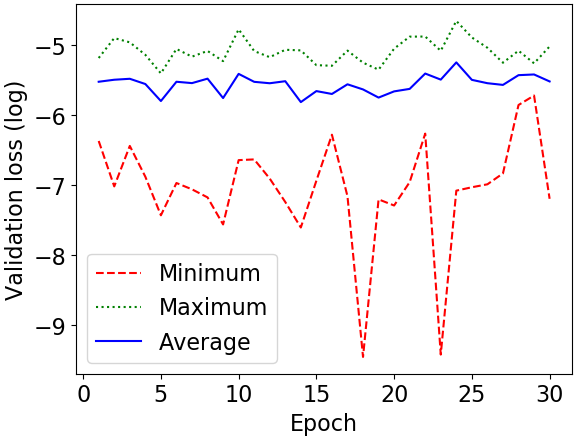}}
    \caption{Progression of the validation loss during  training for LSTM and ReLU models. We show average, minimum and maximum loss. The LSTMs show effective training while the ReLU results mostly stagnate.}
    \label{fig:training-loss}
\end{figure}

Given that low validation losses correlate with high FPF, a possible strategy for improving long term counting behavior could be to train for more epochs to reduce the loss and thus extend the duration of correct counting.   
We show the progression of the validation loss during training for LSTM and ReLU models in Figure~\ref{fig:training-loss}.  
We do not include GRUs as they do not have the capacity for precise counting, so that this strategy does not apply in any case. 

The LSTM training behaves normally, i.e. the  loss value tends to decrease throughout training for the LSTM, but the reduction becomes much slower as the training progresses.
For ReLUs, there is no clear trend, but strong variations throughout. 


In any straightforward solution of exact counting with LSTMs, like the one proposed by \cite{Weiss-et-al-18}, it is necessary to saturate some of the activation functions.
In a cursory inspection we find only a few networks where any saturation occurs.
In all the experimentation we do in the context of this paper, we do not encounter a single LSTM model that actually recognizes any Very Long sequence, even with longer training. 
Thus it seems unlikely that saturation is achievable when training within common ranges. 
Overall, extensive training is not a practical approach to achieve long term counting behavior as it computationally expensive and not reliable. 




\section{Failure Modes}

\begin{figure}[h!]
    \centering
    \subfigure[LSTM - Zigzag 500]{\includegraphics[width=0.325\textwidth]{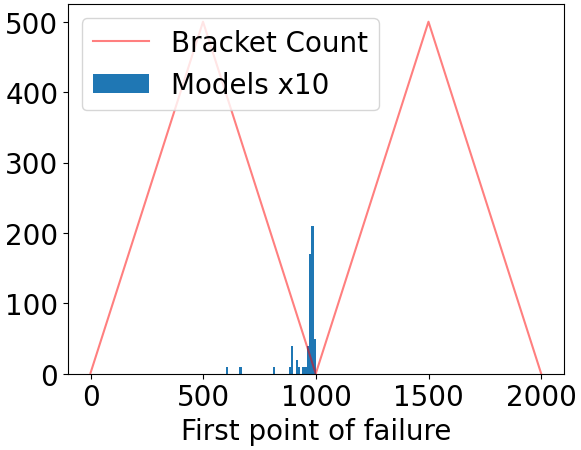}}
    \subfigure[ReLU - Zigzag 500]{\includegraphics[width=0.325\textwidth]{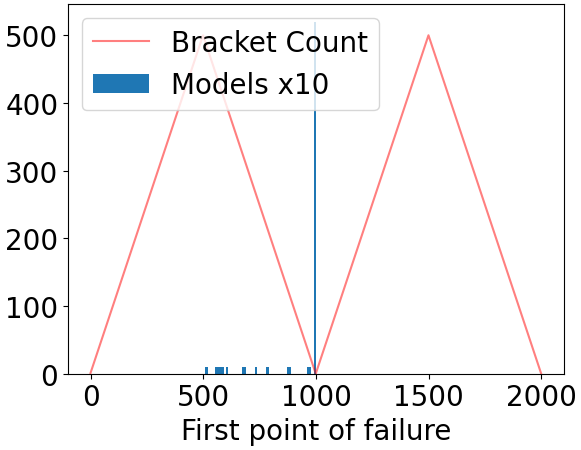}}
    \subfigure[GRU - Zigzag 500]{\includegraphics[width=0.325\textwidth]{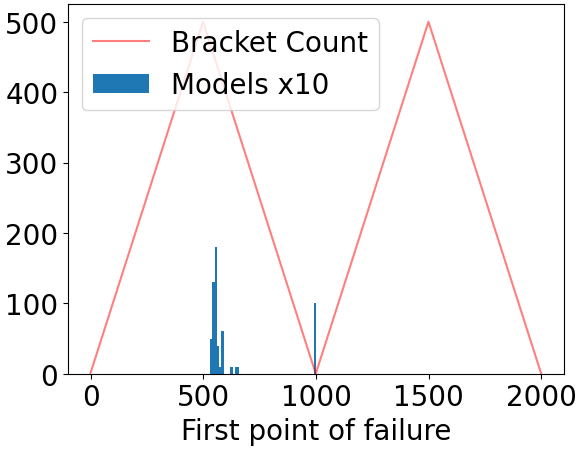}}
    \subfigure[LSTM - Detail]{\includegraphics[width=0.325\textwidth]{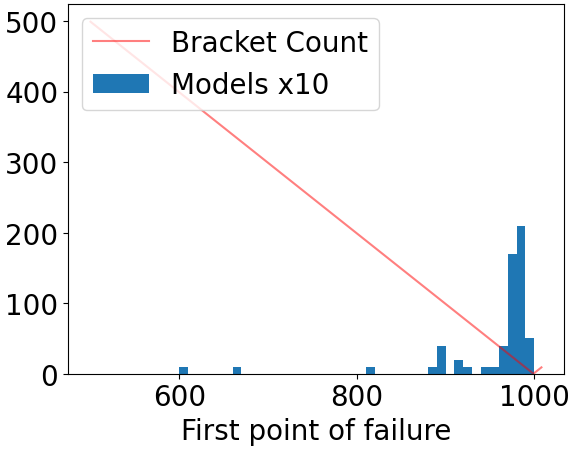}}
    \subfigure[ReLU - Detail]{\includegraphics[width=0.325\textwidth]{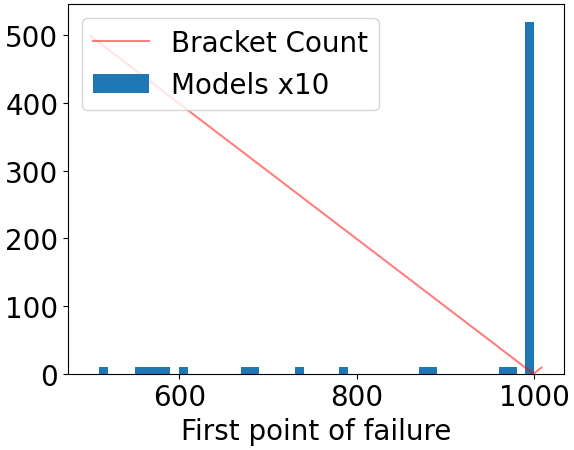}}
    \subfigure[GRU - Detail]{\includegraphics[width=0.325\textwidth]{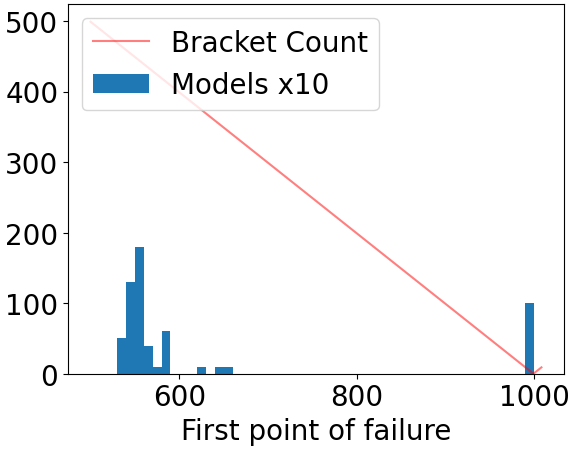}}
    \caption{Histogram of the FPF distribution for the Zigzag sequence with $j = 500$. The red line represents the depth of the sequence (number of opening brackets without a closing bracket), and the histogram represents the number of models failing at different positions. The second row shows details of the distribution from position 500 to 1000, illustrating different failure modes of the different RNN types (see text for a detailed discussion).}
    \label{fig:failure_modes}
\end{figure}


When trying to understand why and how the RNN models fail to count correctly for long sequences, it is worth considering where the counting is represented within the RNN. 
In the single-cell ReLU network, the activation $h_t$ of the ReLU is the only suitable variable. 
For the LSTM, the variable $c_t$ (in the notation used by \citet{Weiss-et-al-18}) is the only one that is not the result of a squashing function, and it has empirically been shown to perform counting in their experiments. 

The only actual test needed of the counter is the comparison to zero to test whether the sequence is a complete Dyck-1 sequence (or $a^n b^n$ in the experiments by \citet{Weiss-et-al-18}). 
In order to achieve zero activation at the appropriate points in the sequence, the increment when processing an opening bracket needs to be of the same absolute value but opposite sign as the decrement when processing a closing bracket. 
The datasets used in these experiments contain only valid Dyck-1 sequences, so that the counter values should not become negative therefore the ReLU should behave linearly. 
For a linear model, the order of the tokens plays no role. 
In the solution provided by \cite{Weiss-et-al-18} for the LSTM it is necessary that several variables saturate the activation functions to reach 1 or -1 for exact counting. 
In particular, if $f_t$ (in their notation) does not saturate, the counter state will be reduced over time, independent of the opening or closing brackets. 
For GRUs, it is not possible to consistently have the same increment and decrement as discussed in \citet{Weiss-et-al-18}. 

We expect these differences in the operation to result in different behaviors that will mainly occur around the points where counter should be 0, because the Dyck-1 sequence is complete. 
For this purpose we create a dataset that consists of repetitions of $j$ opening brackets followed by $j$ closing brackets for $j = \{10, 20, 25, 50, 100, 125, 200, 250, 500, 1000\}$.
All sequences in this dataset are 2000 tokens in length, and distinct.
We refer to this dataset as the \emph{Zigzag Set}.
We test our 60 models for each RNN type at various stages of training, as in the previous section, on the Zigzag Set and record the FPF achieved for each model on each element of the dataset. 
The results on the Zigzag Set are that LSTM and GRU models perform better on the sequences with smaller $j$, showing evidence of a non-linear influence of the counter value (which is proportional to the depth, i.e. the number of open brackets without a corresponding closing bracket).

To illustrate the different behaviors, we plot in  Figure~\ref{fig:failure_modes} the distribution of the FPF values for the Zigzag sequence with $j = 500$ and show in detail graphs positions 500-1000.
At position 1000 we have a complete Dyck-1 sequence, so that the counter should be 0 and the RNN should predict the corresponding class. 
The LSTM models fail mostly at a short distance before position 1000 (Figure~\ref{fig:failure_modes} (a) and (d)), indicating that the decrements are consistently greater by absolute than the increments. 
ReLUs fail mostly at position 1000, indicating that the decrement is less by absolute than the increment, but they also fail before (Figure~\ref{fig:failure_modes} (b) and (e)), indicating that the position of failures is not dependent on the sequence depth. 
The GRUs' increment and decrement values vary throughout the sequence with the decrements relatively large when the counter value is greater. 
This leads to a failure soon after closing bracket sequence starts (Figure~\ref{fig:failure_modes} (c) and (e)). 

The GRU behavior confirms the known limitations.  
The observations of the ReLUs confirm that they can have decrements greater or 
smaller than increments and independent of the sequence depth, but that finding correct weights by backpropagation with gradient descent is difficult. 
LSTMs confirm that exact counting does not occur in a normal training setup and saturation of the activation functions is rarely observed, preventing straightforward solutions. 

This is a potential explanation for the fact that the best model in our experiments was a ReLU model. 
In the ReLU, saturation of an activation function is not possible and not needed, so that it is possible for the learning process to occasionally reach a good weight configuration even if the training process  is not very effective on average.



\section{Conclusions and Future Work}

In this study, we investigate the generalization of RNNs trained to recognize Dyck-1 languages. 
The counting behavior of the trained models is effective on sequences up to double the length of the training sequences, but almost always fails on sequences of 20 times the length (1000 tokens in this case). 
While Training, Validation and Long Test Set losses are predictive for very long term generalization, the correlation is not perfect and training for longer to achieve better generalization is not a practical solution as long training is not efficient on LSTMs and not reliable on ReLUs. 
The different failure patterns of LSTMs, ReLUs and GRUs correspond to our hypotheses of their operation and shows that despite the theoretical capacity of LSTMs and ReLUs, the networks do not reach exact solutions even in the simple noiseless test cases we tried.  

This study empirically extends and explains the theoretical findings by \cite{Weiss-et-al-18} and the experiments by \cite{suzgun2019lstm}.
In the future, there are many interesting research topics stemming form these results.
Understanding the training dynamics of LSTMs and ReLUs better is important, as it will help determine how to initialize and train LSTMs to reach saturation by backpropagation   
and for ReLUs how to learn the correct weight configuration in a stable way. 
Other interesting aspects are the behavior of more commonly used architectures (multi-cell and multi-layer networks), batch training and regularization, and different sequence lengths for training.

Many approaches have been developed to improve the learning of RNNs in general, and they may be helpful in this context. 
E.g., for ReLUs, gradient clipping has been proposed by \citet{DBLP:conf/icml/PascanuMB13} and widely applied.
Specific weight initialization methods may help especially the ReLU networks to learn more effectively \citep{le2015simple}. 
Targeted discretization of model weights or saturating the activation functions may be effective with LSTMs. 
Overall, the learning of discrete counting behavior for the long term with continuous neural networks is not yet solved and more work is needed to understand the theoretical foundations and design effective improvements.


\newpage
\bibliography{anthology,custom}

\begin{thebibliography}{17}
\expandafter\ifx\csname natexlab\endcsname\relax\def\natexlab#1{#1}\fi

\bibitem[{Bengio et~al.(1994)Bengio, Simard, and Frasconi}]{bengio1994learning}
Yoshua Bengio, Patrice Simard, and Paolo Frasconi. 1994.
\newblock Learning long-term dependencies with gradient descent is difficult.
\newblock \emph{IEEE transactions on neural networks}, 5(2):157--166.

\bibitem[{Funahashi and Nakamura(1992)}]{funahashi1992neural}
Ken-Ichi Funahashi and Yuichi Nakamura. 1992.
\newblock \href
  {https://www.kurims.kyoto-u.ac.jp/~kyodo/kokyuroku/contents/pdf/0804-02.pdf}
  {Neural networks, approximation theory, and dynamical systems}.
\newblock \emph{Notes on the Institute of Mathematical Analysis}, 804:18--37.

\bibitem[{Gers and Schmidhuber(2001)}]{DBLP:journals/tnn/GersS01}
Felix~A. Gers and J{\"{u}}rgen Schmidhuber. 2001.
\newblock \href {https://doi.org/10.1109/72.963769} {{LSTM} recurrent networks
  learn simple context-free and context-sensitive languages}.
\newblock \emph{{IEEE} Trans. Neural Networks}, 12(6):1333--1340.

\bibitem[{Hochreiter(1991)}]{hochreiter1991untersuchungen}
Sepp Hochreiter. 1991.
\newblock Untersuchungen zu dynamischen neuronalen netzen.
\newblock \emph{Diploma Thesis, Technische Universit{\"a}t M{\"u}nchen}.

\bibitem[{Hochreiter and Schmidhuber(1997)}]{DBLP:journals/neco/HochreiterS97}
Sepp Hochreiter and J{\"{u}}rgen Schmidhuber. 1997.
\newblock \href {https://doi.org/10.1162/neco.1997.9.8.1735} {Long short-term
  memory}.
\newblock \emph{Neural Comput.}, 9(8):1735--1780.

\bibitem[{Karpathy(2015)}]{karpathy2015unreasonable}
Andrej Karpathy. 2015.
\newblock \href {http://karpathy.github.io/2015/05/21/rnn-effectiveness/} {The
  unreasonable effectiveness of recurrent neural networks}.
\newblock \emph{Andrej Karpathy blog}, 21:23.

\bibitem[{Kingma and Ba(2014)}]{kingma2014adam}
Diederik~P. Kingma and Jimmy Ba. 2014.
\newblock Adam: A method for stochastic optimization.
\newblock \emph{arXiv preprint arXiv:1412.6980}.

\bibitem[{Le et~al.(2015)Le, Jaitly, and Hinton}]{le2015simple}
Quoc~V. Le, Navdeep Jaitly, and Geoffrey~E. Hinton. 2015.
\newblock A simple way to initialize recurrent networks of rectified linear
  units.
\newblock \emph{arXiv preprint arXiv:1504.00941}.

\bibitem[{Leshno et~al.(1993)Leshno, Lin, Pinkus, and
  Schocken}]{leshno1993multilayer}
Moshe Leshno, Vladimir~Ya Lin, Allan Pinkus, and Shimon Schocken. 1993.
\newblock Multilayer feedforward networks with a nonpolynomial activation
  function can approximate any function.
\newblock \emph{Neural networks}, 6(6):861--867.

\bibitem[{Merrill(2020)}]{DBLP:journals/corr/abs-2004-06866}
William Merrill. 2020.
\newblock \href {http://arxiv.org/abs/2004.06866} {On the linguistic capacity
  of real-time counter automata}.
\newblock \emph{CoRR}, abs/2004.06866.

\bibitem[{Merrill et~al.(2020)Merrill, Weiss, Goldberg, Schwartz, Smith, and
  Yahav}]{DBLP:conf/acl/MerrillWGSSY20}
William Merrill, Gail Weiss, Yoav Goldberg, Roy Schwartz, Noah~A. Smith, and
  Eran Yahav. 2020.
\newblock \href {https://doi.org/10.18653/v1/2020.acl-main.43} {A formal
  hierarchy of {RNN} architectures}.
\newblock In \emph{Proceedings of the 58th Annual Meeting of the Association
  for Computational Linguistics, {ACL} 2020, Online, July 5-10, 2020}, pages
  443--459. Association for Computational Linguistics.

\bibitem[{Pascanu et~al.(2013)Pascanu, Mikolov, and
  Bengio}]{DBLP:conf/icml/PascanuMB13}
Razvan Pascanu, Tom{\'{a}}s Mikolov, and Yoshua Bengio. 2013.
\newblock \href {http://proceedings.mlr.press/v28/pascanu13.html} {On the
  difficulty of training recurrent neural networks}.
\newblock In \emph{Proceedings of the 30th International Conference on Machine
  Learning, {ICML} 2013, Atlanta, GA, USA, 16-21 June 2013}, volume~28 of
  \emph{{JMLR} Workshop and Conference Proceedings}, pages 1310--1318.
  JMLR.org.

\bibitem[{Siegelmann and Sontag(1995)}]{DBLP:journals/jcss/SiegelmannS95}
Hava~T. Siegelmann and Eduardo~D. Sontag. 1995.
\newblock \href {https://doi.org/10.1006/jcss.1995.1013} {On the computational
  power of neural nets}.
\newblock \emph{J. Comput. Syst. Sci.}, 50(1):132--150.

\bibitem[{Srivastava et~al.(2022)Srivastava, Rastogi, Rao, Shoeb, Abid, Fisch,
  Brown, Santoro, Gupta, Garriga-Alonso et~al.}]{srivastava2022beyond}
Aarohi Srivastava, Abhinav Rastogi, Abhishek Rao, Abu Awal~Md Shoeb, Abubakar
  Abid, Adam Fisch, Adam~R Brown, Adam Santoro, Aditya Gupta, Adri{\`a}
  Garriga-Alonso, et~al. 2022.
\newblock Beyond the imitation game: Quantifying and extrapolating the
  capabilities of language models.
\newblock \emph{arXiv preprint arXiv:2206.04615}.

\bibitem[{Suzgun et~al.(2019)Suzgun, Gehrmann, Belinkov, and
  Shieber}]{suzgun2019lstm}
Mirac Suzgun, Sebastian Gehrmann, Yonatan Belinkov, and Stuart~M Shieber. 2019.
\newblock \href {https://aclanthology.org/W19-3905.pdf} {Lstm networks can
  perform dynamic counting}.
\newblock \emph{ACL 2019}, page~44.

\bibitem[{Vaswani et~al.(2017)Vaswani, Shazeer, Parmar, Uszkoreit, Jones,
  Gomez, Kaiser, and Polosukhin}]{vaswani2017attention}
Ashish Vaswani, Noam Shazeer, Niki Parmar, Jakob Uszkoreit, Llion Jones,
  Aidan~N Gomez, {\L}ukasz Kaiser, and Illia Polosukhin. 2017.
\newblock Attention is all you need.
\newblock In \emph{Advances in neural information processing systems}, pages
  5998--6008.

\bibitem[{Weiss et~al.(2018)Weiss, Goldberg, and Yahav}]{Weiss-et-al-18}
Gail Weiss, Yoav Goldberg, and Eran Yahav. 2018.
\newblock \href {https://doi.org/10.18653/v1/P18-2117} {On the practical
  computational power of finite precision rnns for language recognition}.
\newblock In \emph{Proceedings of the 56th Annual Meeting of the Association
  for Computational Linguistics, {ACL} 2018, Melbourne, Australia, July 15-20,
  2018, Volume 2: Short Papers}, pages 740--745. Association for Computational
  Linguistics.

\end{thebibliography}
\bibliographystyle{acl_natbib}

\end{document}